%% file: interspeech_cameraready.tex
\documentclass{INTERSPEECH2023}
\usepackage[utf8]{inputenc}
\usepackage[]{geometry}
\usepackage{setspace}
\usepackage{listings}
\usepackage{graphicx}
\usepackage{subcaption}
\usepackage{cleveref}
\usepackage{pifont}
\usepackage{multirow}
\usepackage{tablefootnote}

\usepackage{float}

\interspeechcameraready

\usepackage{todonotes}
\usepackage{booktabs}
\title{Wave to Syntax: Probing spoken language models for syntax}
\name{Gaofei Shen$^1$, Afra Alishahi$^1$, Arianna Bisazza$^2$, Grzegorz Chrupa\l{}a$^1$}
\address{
  $^1$Tilburg University, the Netherlands\\
  $^2$University of Groningen, the Netherlands}
\email{g.shen@tilburguniversity.edu, a.alishahi@uvt.nl, a.bisazza@rug.nl, grzegorz@chrupala.me}

\begin{document}

\maketitle
 
\begin{abstract}
Understanding which information is encoded in deep models of spoken and written language has been the focus of much research in recent years, as it is crucial for debugging and improving these architectures. Most previous work has focused on probing for speaker characteristics, acoustic and phonological information in models of spoken language, and for syntactic information in models of written language. Here we focus on the encoding of syntax in several self-supervised and visually grounded models of spoken language. We employ two complementary probing methods, combined with baselines and reference representations to quantify the degree to which syntactic structure is encoded in the activations of the target models. We show that syntax is captured most prominently in the middle layers of the networks, and more explicitly within models with more parameters.\footnote{Code: \url{https://github.com/techsword/wave-to-syntax}}
\end{abstract}

\input{introduction}

\input{related}

\input{methods}

\input{results}

\section{Conclusions}

We use two established probing techniques to assess the 
amount of syntactic information encoded by several spoken 
language models. The results from both probes confirm that 
spoken language models encode a moderate level of syntactic 
information. We see that different training objectives 
considerably affect the degree of syntax encoded in each layer of the models and
so does model size, with text-based training and larger 
model size leading to higher syntactic probe scores.

Our study only looks at sentence-level representations: 
it would also be interesting to extend our
experiments to sub-sentence level probing \cite{Hewitt2019ASP}. 
Additionally, while we only
use English datasets in this work, future studies could 
compare the ability of large-scale spoken language models
to encode syntactic structures across different languages.

\section{Acknowledgements}
This publication is part of the project \textit{InDeep: Interpreting Deep Learning Models for Text and Sound} (with project number NWA.1292.19.399) of the National Research Agenda (NWA-ORC) program.

\bibliographystyle{IEEEtran}
\bibliography{w2s}

\end{document}

%% file: introduction.tex
\section{Introduction}
\label{sec:intro}

State-of-the-art models of (spoken) language rely on deep learning
architectures composed of various components and based especially on
the Transformer model (e.g.\ \cite{baevskiWav2vecFrameworkSelfSupervised2020, hsuHuBERTSelfSupervisedSpeech2021}). Evaluating the performance of these models via
standard quantitative protocols is straightforward enough; but it
is not always easy to understand the reasons for fine-grained patterns
of behavior and failure modes, and not trivial to debug and iterate on
design. One tool to aid in this process has been the analysis and
interpretation of the representations learned by the models, as
encoded in the activation patterns within the various components \cite{Belinkov2018AnalysisMI, Rogers2020API}.

For text-based models, numerous works have probed these
activations for many types of information, with a special interest in
syntactic structure \cite{conneauWhatYouCan2018, Hewitt2019ASP, chrupalaCorrelatingNeuralSymbolic2019}.
By contrast, the focus on speech models has been on the encoding of acoustic information,
speaker characteristics, phonetics and phonology (e.g.\ \cite{Krug2018NeuronAP, chrupalaAnalyzingAnalyticalMethods2020, pasadLayerwiseAnalysisSelfsupervised2021}).

However, as the current crop of speech models grows in size and
sophistication, we ask whether they also learn to encode
syntactic structure to any appreciable degree. If the knowledge of
syntax is useful for optimizing a model's training objective, the
expectation would be that, given enough data, the model will learn to
represent it. As a simple example, consider the utterance \textit{The
  \underline{authors} of the book \underline{are} French}: if the
timesteps in the feature space corresponding to \textit{are} are
masked and the model needs to reconstruct them, then it can do it
better if it encodes number agreement between subject and
verb. In the current study, we evaluate the hypothesis that syntax
information is in general represented in such models.

We use two established representation probing techniques 
\cite{conneauWhatYouCan2018, chrupalaCorrelatingNeuralSymbolic2019} in
combination with carefully designed baselines and reference
representations to quantify the encoding of syntactic
structure in selected current models of spoken language trained
via self-supervision objectives. We also apply our methodology to
models trained via visual supervision (also known as visual grounding)
and to a model trained via a combination of self- and visual
supervision. We track the encoding of syntactic structure throughout
the transformer layers of these target models.

Our findings show
that syntax is captured by all these models, with the following
caveats and details: Firstly, the encoding of syntax is generally
weaker than in text-based models (such as BERT \cite{devlinBERTPretrainingDeep2019})
. Secondly,
much of the syntactic structure that is captured may be encoded in
lexical rather than purely syntactic form. Thirdly, self-supervised
and combined objectives lead to less syntactic encoding in
the final model layers, while the visually-supervised objective
does not have this effect. Finally, increased model size is associated
with the stronger encoding of syntactic information.

%% file: related.tex
\section{Related work}
\label{sec:related}

Within Natural Language Processing (NLP), there has been substantial
interest in understanding the representations emerging within text-based 
language models: surveys of such work include
\cite{Belinkov2018AnalysisMI,Rogers2020API,Sajjad2021NeuronlevelIO}. 
The predominant family of approaches relies on correlating the activation patterns in trained models
to linguistic structures that are considered necessary for correctly processing natural language. 
For written language, these are often various types of word categories,
constituency structures or syntactic and semantic dependencies, for
example as proposed in
\cite{conneauWhatYouCan2018,Hewitt2019ASP,chrupalaCorrelatingNeuralSymbolic2019}.

For models of spoken language, most previous work has focused on
acoustic, phonetic and phonemic structures as well
as on speaker characteristics, which are the most plausible types
of information a speech model is expected to learn. Works
analyzing the encoding of phonemes in a variety of speech models
ranging from basic CNN-based ASR models to current transformer-based
self-supervised models
\cite{Krug2018NeuronAP,chrupalaAnalyzingAnalyticalMethods2020,Nagamine2015ExploringHD,Belinkov2017AnalyzingHR,Belinkov2019AnalyzingPA,deSeyssel2022ProbingPL} tend to find a salient encoding of phonemes in
some layers of the models analyzed. In
\cite{pasadLayerwiseAnalysisSelfsupervised2021} the authors check for
the encoding of acoustic, phonemic, lexical and semantic information
in the self-supervised
wav2vec2~\cite{baevskiWav2vecFrameworkSelfSupervised2020}
using probes such as
canonical correlation analysis and mutual information, finding an
autoencoder-style behavior, where across the layers the
representations first diverge from low-level input features and at the
end approximate them again. The reverse is the case for higher-level
information such as word identity and meaning. 
A few papers have focused on analyzing phonology and/or semantics in
visually-grounded models \cite{Chrupaa2021VisuallyGM}, for example
\cite{alishahiEncodingPhonologyRecurrent2017,Khorrami2021CanPS}, in
general finding phonemes encoded in lower layers and semantics in higher
layers. 

Much less work has looked at syntactic structures in models of spoken
language: one partial exception is \cite{Singla2022WhatDA}, who probe 
wav2vec2 and Mockingjay~\cite{Liu2019MockingjayUS} for the encoding of
acoustic and linguistic information, including syntax tree depth. 
The results reported are quite surprising and even
implausible in that the encoding of most linguistic
features in the speech models is found to be stronger than in the text-based
BERT. 

In this paper, we focus exclusively on probing for syntactic
structures, aiming to examine them in several large-scale speech
models, while taking care to include two independent methods as well
as all the appropriate baselines and sanity checks in order to
quantify the encoding of syntax in a reliable way.

%% file: methods.tex
\section{Methods}
This study aims to reliably establish the presence of
syntactic information in models of spoken language. We, therefore, use
established methods for this type of analysis and focus on careful
experimental design rather than technical novelty. We use two
separate probing techniques \cite{conneauWhatYouCan2018, chrupalaCorrelatingNeuralSymbolic2019}, with several target models trained and
tested on two different datasets.

\subsection{Datasets}
We use two English audio datasets for the current study:
LibriSpeech~\cite{panayotovLibrispeechASRCorpus2015}
and SpokenCOCO~\cite{hsuTextFreeImagetoSpeechSynthesis2021}. LibriSpeech consists of
audiobook recordings from the LibriVox project with a total of 960
hours of audio. SpokenCOCO is a spoken version of the image caption
dataset COCO~\cite{linMicrosoftCOCOCommon2015} with more than
600,000 spoken utterances paired with text captions and images. We use the LibriSpeech 
train-100h split and the SpokenCOCO validation split in our probing experiment to reduce computational load. 
We filter out utterances longer than 52 words for LibriSpeech and those longer than 20 words for SpokenCOCO.
\Cref{tab:data} shows the details of the data used in our experiments.

\begin{table}[h]
    \caption[short]{Datasets used for this study.} \label[table]{tab:data}
    \centering
    \begin{tabular}{ l r r}
        \toprule
        Name  & \#Utts. & \# Filtered Utts.\\
        \midrule
        LibriSpeech  & 24,766 & 24,592\\
        SpokenCOCO  & 28,539 & 27,496\\
        \bottomrule
    \end{tabular}
\end{table}

\subsection{Target Models}

For this study, we use the following model variants as specified in \Cref{tab:models}:
\begin{description}
\item[wav2vec2] \cite{baevskiWav2vecFrameworkSelfSupervised2020} is
  pre-trained on LibriSpeech to discriminate masked time steps in feature encoder outputs. 
  In addition to the base pre-trained model,
  we also include a fine-tuned base model and 
  a fine-tuned large model. The large model\footnote{ \url{https://huggingface.co/jonatasgrosman/wav2vec2-large-english}} 
  is fine-tuned for English ASR on the same dataset. Additionally, we 
  fine-tuned the base model for English ASR on our experimental data for 
  10,000 steps, which enables us to check the effect of model size.
\item[HuBERT] \cite{hsuHuBERTSelfSupervisedSpeech2021} is similar to
  the wav2vec2 architecture but pre-trained on labels created off-line via
  clustering; also pre-trained on LibriSpeech.
\item[FaST-VGS] \cite{pengSelfSupervisedRepresentationLearning2022} is
  a visually grounded model based on the wav2vec2 architecture. It loads 
  model weights from the pre-trained wav2vec2-base with randomly reinitialized
  final four layers and is further trained 
  on SpokenCOCO to match images with the speech that describes them.
\item[FaST-VGS+] \cite{pengSelfSupervisedRepresentationLearning2022}
  same as the previous model, but trained on both SpokenCOCO and
  LibriSpeech with a combination of the visually-grounded loss and the
  wav2vec2 self-supervised loss.
\item[BERT] \cite{devlinBERTPretrainingDeep2019}
  text-based language model included as a ceiling reference. 
  BERT is pre-trained on 3,300M words of books and web content.

\item[BoW] a text-based bag-of-words representation constructed from the combined 
  text from both datasets with all sentences containing non-Latin 
  characters removed. This representation captures all the words in the utterance, 
  but no word-order information. 
\end{description}

\begin{table}[h]
  \centering
  \caption{\label[table]{tab:models} Models investigated in this
    study. PT = Pre-Trained, FT = Fine-Tuned,
    SS = Self-Supervised, VS = Visually Supervised, AV = Audio-Visual.}
  \begin{tabular}{ l l l l l}
    \toprule
      Model         & Size & Train.  & Loss & Mod. \\
      \midrule
      wav2vec2      & base  & PT        & SS   & Audio\\
      wav2vec2      & base  & FT        & SS   & Audio\\
      wav2vec2      & \underline{large}\tablefootnote{The base models have 12 transformer layers and a hidden size of 768; 
      the large model has 24 layers and a hidden size of 1024.} 
                            & FT        & SS   & Audio\\
      HuBERT        & base  & PT        & SS   & Audio\\
      FaST-VGS      & base   & PT        & VS   & AV\\
      FaST-VGS+     & base   & PT        & SS+VS & AV\\
      \midrule
      BERT          & base  & PT        & SS   & Text\\
      \midrule
      BoW           &       &           &      & Text \\
      \bottomrule
  \end{tabular}
\end{table}

\subsection{TreeDepth Probe}
\label[section]{sec:treedepth}
The objective of this probe is to predict the maximum depth of the
constituency tree of a given utterance from the activation pattern in each transformer layer
of a model when processing this utterance.

We generated hidden-state outputs and applied mean-pooling along the
time axis to generate utterance vectors for all transformer layers of
each model. We used the Stanza parser \cite{qiStanzaPythonNatural2020} to
generate constituency trees for all utterances and calculate their tree depth. 
We fit a ridge regression model on
embeddings generated for both LibriSpeech and SpokenCOCO. As
controls, wordcount, and bag-of-words (BoW) model
representation and their combinations with the embeddings were used in
training the regression model as well. A 75:25 train-test split was
used, and the model selected via 10-fold cross-validation was then evaluated
on the test split and its score reported.

\subsection{TreeKernel Probe}
\label[section]{sec:treekernel}

Representational Similarity Analysis (RSA) \cite{kriegeskorte2008representational} is a method which correlates the
similarity structures of two representational spaces. 
RSA\textsubscript{regress}~\cite{chrupalaCorrelatingNeuralSymbolic2019} introduces a trainable version of RSA,
where two input and output spaces are set up in terms of vectors of
similarity to a held-out set of anchor points, and then
a multivariate regression model is fit to map between
them. 

As in the original paper, we used cosine
similarity between vectors in the input space as metric, and tree kernels between pairs of
syntax trees in the output space. Tree kernel is a
way of measuring similarities between syntactic trees by
efficiently computing the proportion of shared tree segments. After obtaining the
constituency trees, we delexicalized the trees to ensure the tree
kernel is only based on structure and not word overlap. We used the algorithms
introduced in
\cite{moschittiMakingTreeKernels2006,collinsConvolutionKernelsNatural2001}
for computing the normalized tree kernel; we closely followed the
specific details in \cite{chrupalaCorrelatingNeuralSymbolic2019}, using parameter
$\lambda = \frac{1}{2}$ and two hundred anchor sentences. The score of
the model selected via 10-fold cross-validation is reported.

For both probes, the hyperparameter tuned was regularization strength
$\alpha$ for values $\{ 10^n~|~ n \in \{-3, -2, -1, 0, 1, 2\} \}$.

%% file: results.tex
\section{Results}
\begin{figure}[ht]
    \captionsetup[subfigure]{aboveskip=-0.5pt,belowskip=-0.5pt}

    \centering
    \begin{subfigure}[b]{\linewidth}

    \includegraphics[width = \linewidth]{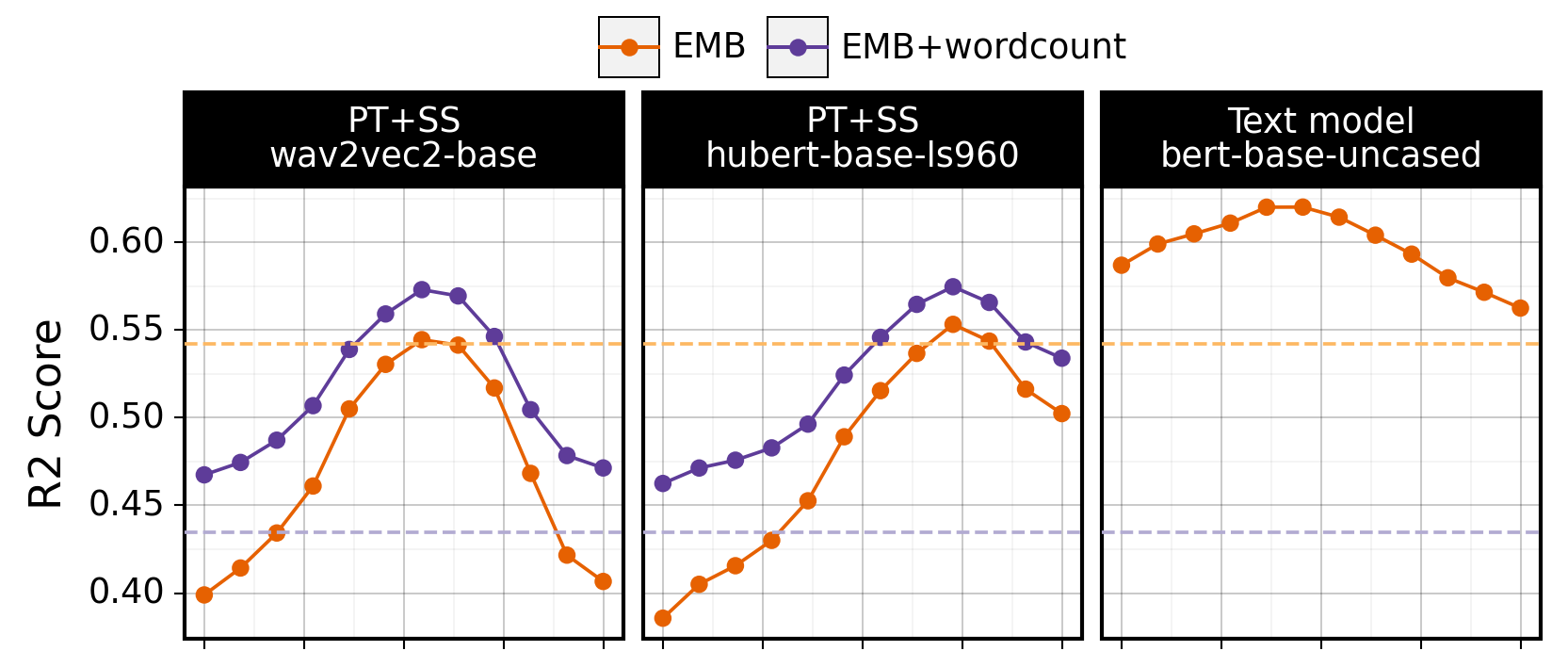}
    \caption{Text vs.\ audio models \label[figure]{fig:text_vs_pt}}
    \end{subfigure}
    \begin{subfigure}[b]{\linewidth}

        \includegraphics[width = \linewidth]{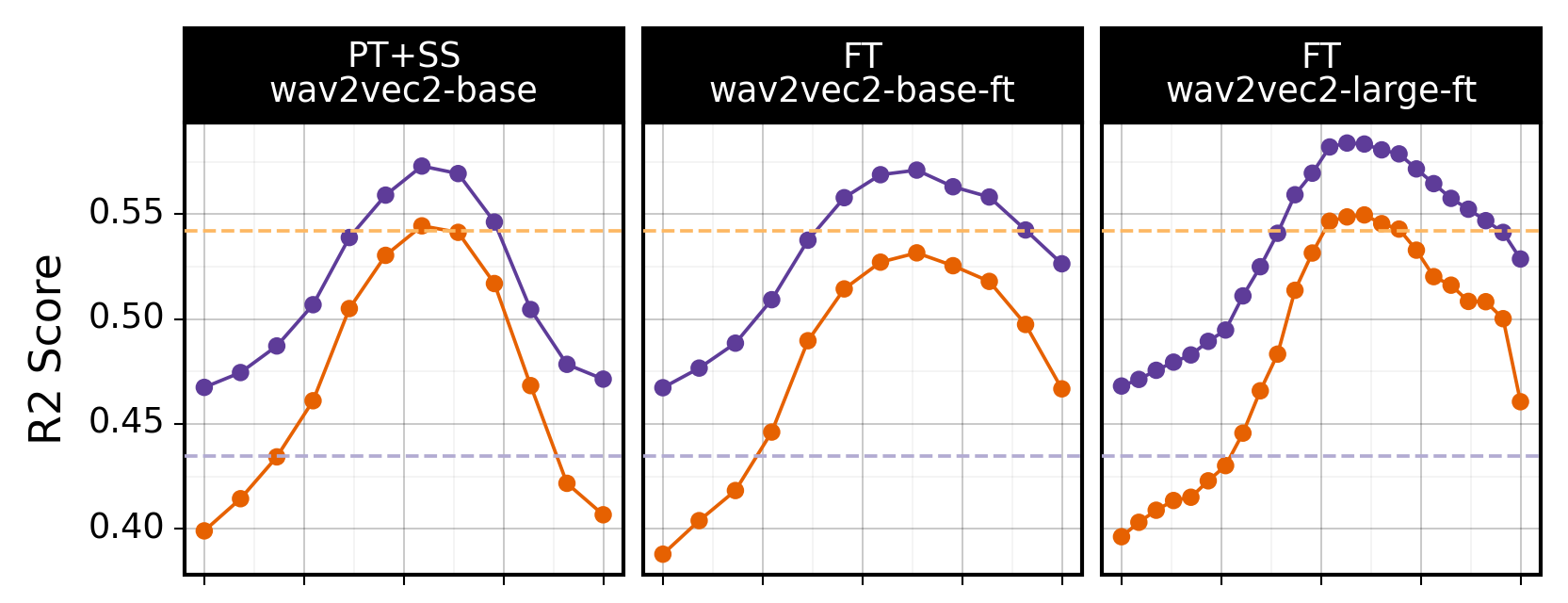}
        \caption{Pre-training vs.\ fine-tuning \label[figure]{fig:ptvsft_td}}
    \end{subfigure}
    \begin{subfigure}[b]{\linewidth}

        \includegraphics[width = \linewidth]{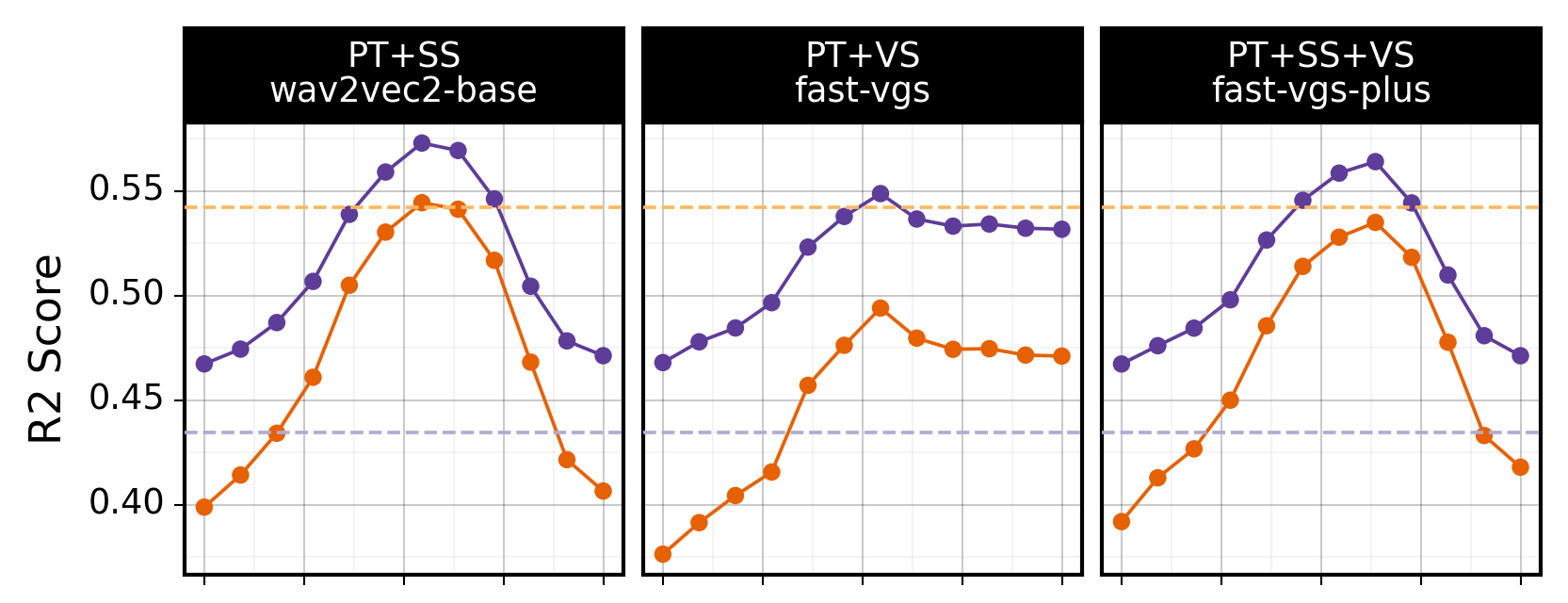}
        \caption{Self vs.\ visual supervision \label[figure]{fig:avsv_td}}
    \end{subfigure}

    \caption{$R^2$ scores for predicting TreeDepth from
      embeddings (Librispeech). X-axis = transformer layer from shallow to deep, orange
      dashed line = BoW reference, purple dashed line = Word count reference. See \Cref{tab:models} for panel headings. \label[figure]{fig:treedepth}}

\end{figure}

\begin{figure}[ht]
    \captionsetup[subfigure]{aboveskip=-0.5pt,belowskip=-0.5pt}
    \centering
    \begin{subfigure}[b]{\linewidth}

        \includegraphics[width = \linewidth]{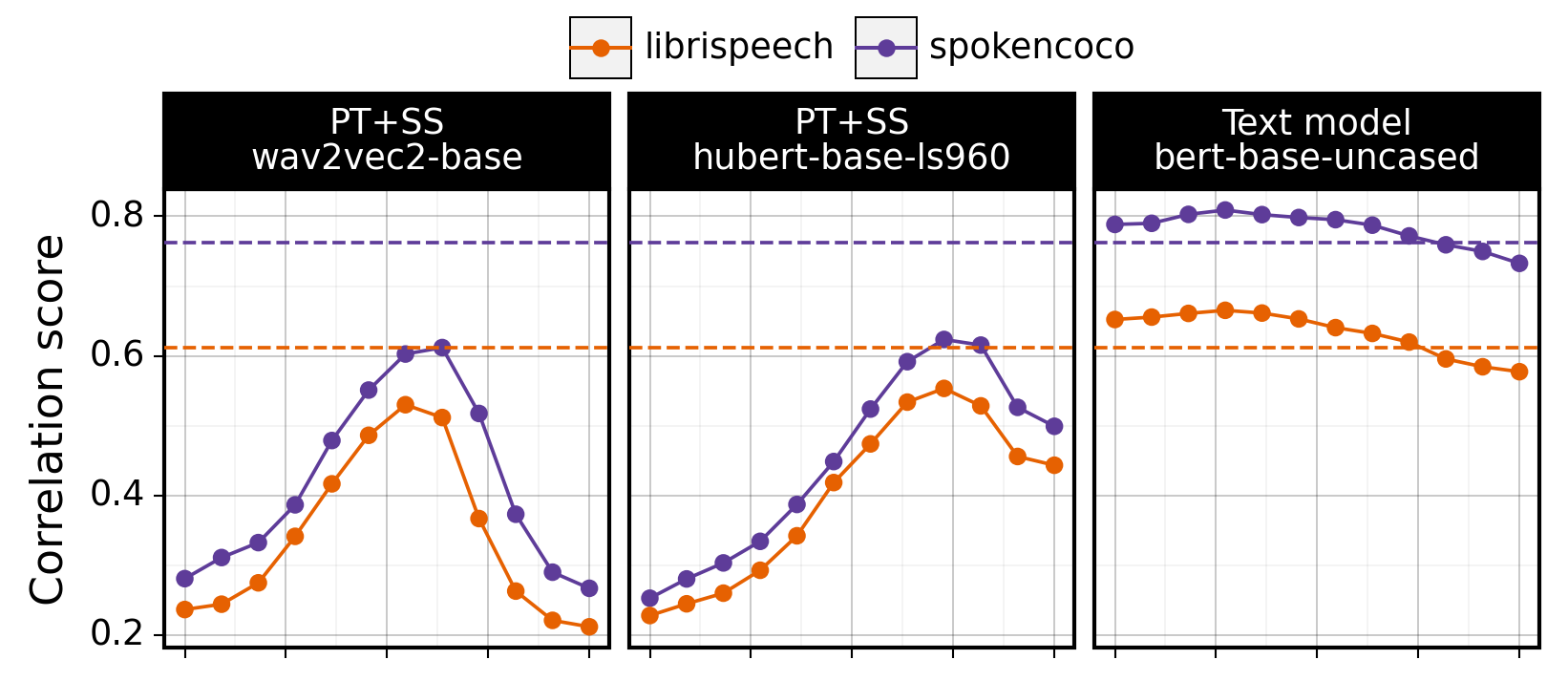}
        \caption{Text vs.\ audio models \label[figure]{fig:textvspt_tk}}
    \end{subfigure}
    \begin{subfigure}[b]{\linewidth}

    \includegraphics[width = \linewidth]{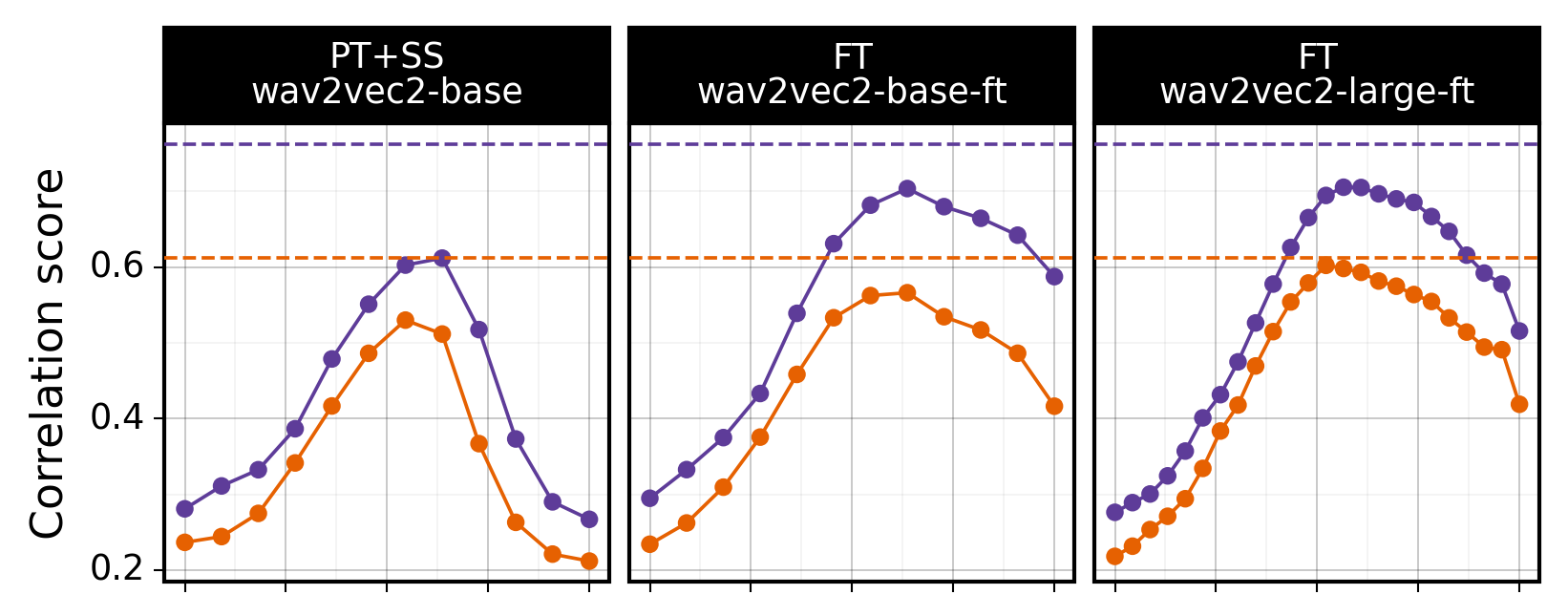}
    \caption{Pre-training vs.\ fine-tuning \label[figure]{fig:ptvsft_tk}}
    \end{subfigure}
    \begin{subfigure}[b]{\linewidth}

        \includegraphics[width = \linewidth]{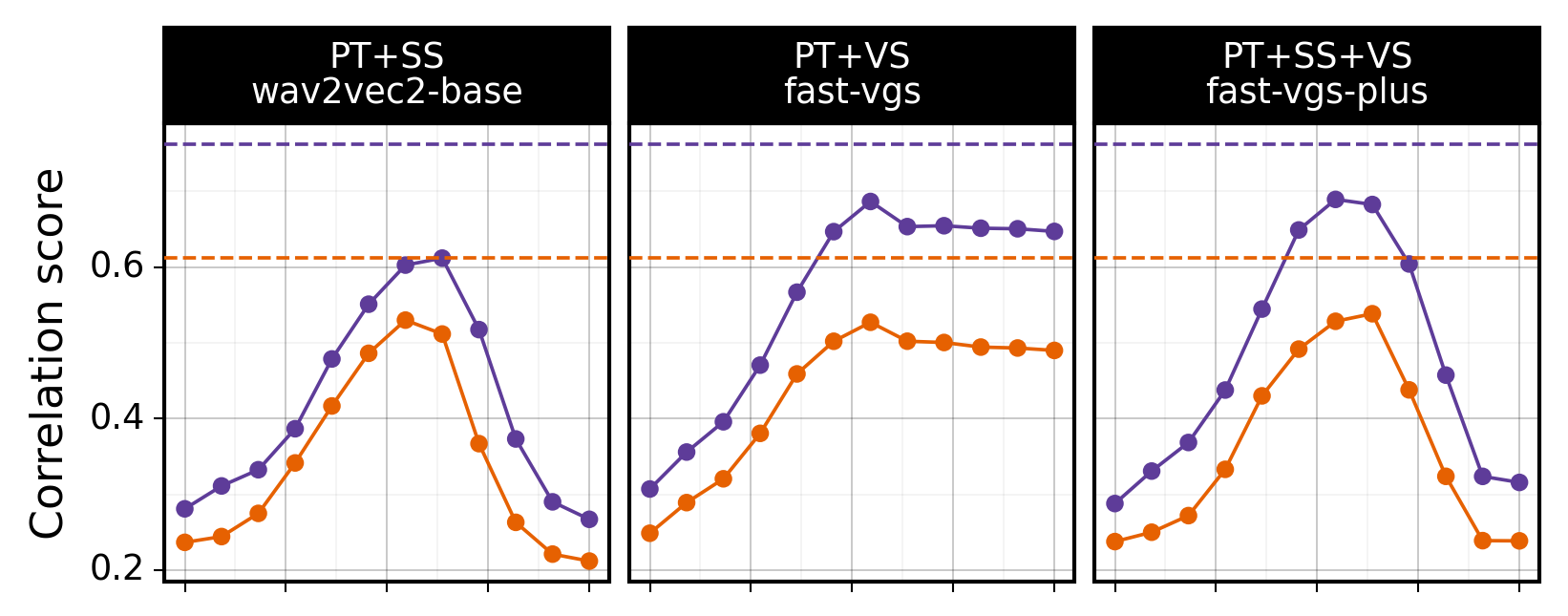}
        \caption{Self vs.\ visual supervision \label[figure]{fig:avsv_tk}}
    \end{subfigure}
    \caption{$R^2$ scores for embedding distances and
      TreeKernel. X-axis = transformer layer from shallow to
      deep. Dashed lines = BoW reference. See \Cref{tab:models} for panel headings. \label[figure]{fig:treekernel}}

\end{figure}

\Cref{fig:treedepth,fig:treekernel} show the results for the TreeDepth
and TreeKernel probing tasks, respectively. For clarity, \Cref{fig:treedepth}
only shows results on Librispeech.

\paragraph*{Syntax encoding in spoken vs.\ text models.}
\Cref*{fig:text_vs_pt} shows that spoken language models 
feature the strongest encoding of tree depth in the middle to deep layers. Concatenating 
wordcount features to embeddings in the TreeDepth task improves the probing results in the 
middle layers and outperforms the BoW reference. 
\Cref*{fig:textvspt_tk} shows a similar trend for the \mbox{TreeKernel} probe.

As expected, both probing tasks show that syntactic information 
is encoded more strongly and consistently in text models across all
layers.\footnote{Our results contradict those of
  \cite{Singla2022WhatDA}: possibly due to them
report probing scores directly on probe-training data: the description
in the paper is unclear on this point.}
However, TreeKernel results for the middle layers of the spoken language models 
come close to the text models, with wav2vec2-large-ft having the highest 
correlation score. Overall, these results suggest that spoken 
language models encode syntactic structure to a moderate degree.
Comparison with the BoW reference suggests that most of the syntactic
information encoded in speech models is entangled with lexical
representations, rather than being abstract.

\paragraph*{Pre-training vs.\ fine-tuning}
\Cref{fig:ptvsft_td} shows that models fine-tuned on ASR achieve
higher scores in the TreeDepth probe than the pre-trained models. The
pre-trained models have a large dip at the final layers with the
scores go back to those for the first layer.  For the fine-tuned
models, the final layer dip is less pronounced.  The pre-training
objective is to reconstruct/discriminate audio features, whereas the fine-tuning objective is to
output well-formed transcriptions, which
requires more syntactic information to be encoded in the activation
patterns, resulting in the contrast between the final layer scores.

Comparing the panels in \Cref{fig:ptvsft_tk}, we can see that while
fine-tuning increases the amount of syntactic information encoded by
the model, the size of the model also matters. The larger hidden size
and deeper model architecture prove useful in encoding extra
information. Wav2vec2-large-ft was also fine-tuned on significantly more
data than wav2vec2-base-ft.

\paragraph*{Self- vs. Visual supervision}
As shown in \Cref{fig:avsv_td,fig:avsv_tk}, the score curve from
FaST-VGS+ has a similar shape as wav2vec2-base. In addition to
visual supervision, the plus variant
also uses the same masked language modeling loss as wav2vec2, and therefore
the two models behave similarly. In comparison, the final layer
dip is absent for the FaST-VGS model, likely due to the fact
that FaST-VGS does not use the self-supervised objective, and thus
not display a decrease in syntax encoding in the final layers.

%% file: interspeech_cameraready.bbl
\begin{thebibliography}{10}
\providecommand{\url}[1]{#1}
\csname url@samestyle\endcsname
\providecommand{\newblock}{\relax}
\providecommand{\bibinfo}[2]{#2}
\providecommand{\BIBentrySTDinterwordspacing}{\spaceskip=0pt\relax}
\providecommand{\BIBentryALTinterwordstretchfactor}{4}
\providecommand{\BIBentryALTinterwordspacing}{\spaceskip=\fontdimen2\font plus
\BIBentryALTinterwordstretchfactor\fontdimen3\font minus
  \fontdimen4\font\relax}
\providecommand{\BIBforeignlanguage}[2]{{%
\expandafter\ifx\csname l@#1\endcsname\relax
\typeout{** WARNING: IEEEtran.bst: No hyphenation pattern has been}%
\typeout{** loaded for the language `#1'. Using the pattern for}%
\typeout{** the default language instead.}%
\else
\language=\csname l@#1\endcsname
\fi
#2}}
\providecommand{\BIBdecl}{\relax}
\BIBdecl

\bibitem{baevskiWav2vecFrameworkSelfSupervised2020}
A.~Baevski, H.~Zhou, A.~Mohamed, and M.~Auli, ``Wav2vec 2.0: A framework for
  self-supervised learning of speech representations,'' in \emph{NIPS 2020},
  2020.

\bibitem{hsuHuBERTSelfSupervisedSpeech2021}
W.-N. Hsu, B.~Bolte, Y.-H.~H. Tsai, K.~Lakhotia, R.~Salakhutdinov, and
  A.~Mohamed, ``Hubert: Self-supervised speech representation learning by
  masked prediction of hidden units,'' \emph{TASLP}, vol.~29, pp. 3451--3460,
  2021.

\bibitem{Belinkov2018AnalysisMI}
Y.~Belinkov and J.~R. Glass, ``Analysis methods in neural language processing:
  A survey,'' \emph{TACL}, vol.~7, pp. 49--72, 2018.

\bibitem{Rogers2020API}
A.~Rogers, O.~Kovaleva, and A.~Rumshisky, ``A primer in bertology: What we know
  about how bert works,'' \emph{TACL}, vol.~8, pp. 842--866, 2020.

\bibitem{conneauWhatYouCan2018}
A.~Conneau, G.~Kruszewski, G.~Lample, L.~Barrault, and M.~Baroni, ``What you
  can cram into a single {\$}{\&}!{\#}* vector: Probing sentence embeddings for
  linguistic properties,'' in \emph{ACL 2018}, vol. 1: Long Papers, Jul. 2018,
  pp. 2126--2136.

\bibitem{Hewitt2019ASP}
J.~Hewitt and C.~D. Manning, ``A structural probe for finding syntax in word
  representations,'' in \emph{NAACL}, 2019.

\bibitem{chrupalaCorrelatingNeuralSymbolic2019}
\BIBentryALTinterwordspacing
G.~Chrupała and A.~Alishahi, ``Correlating {{Neural}} and {{Symbolic
  Representations}} of {{Language}},'' in \emph{ACL 2019}.\hskip 1em plus 0.5em
  minus 0.4em\relax {ACL}, 2019, pp. 2952--2962. [Online]. Available:
  \url{https://aclanthology.org/P19-1283}
\BIBentrySTDinterwordspacing

\bibitem{Krug2018NeuronAP}
A.~Krug, R.~Knaebel, and S.~Stober, ``Neuron activation profiles for
  interpreting convolutional speech recognition models,'' in
  \emph{NeurIPS-IRASL}, 2018.

\bibitem{chrupalaAnalyzingAnalyticalMethods2020}
G.~Chrupała, B.~Higy, and A.~Alishahi, ``Analyzing analytical methods: {{The}}
  case of phonology in neural models of spoken language,'' in \emph{ACL
  2020}.\hskip 1em plus 0.5em minus 0.4em\relax {ACL}, 2020, pp. 4146--4156.

\bibitem{pasadLayerwiseAnalysisSelfsupervised2021}
A.~Pasad, J.-C. Chou, and K.~Livescu, ``Layer-wise analysis of a
  self-supervised speech representation model,'' in \emph{IEEE ASRU 2021},
  2021, pp. 914--921.

\bibitem{devlinBERTPretrainingDeep2019}
\BIBentryALTinterwordspacing
J.~Devlin, M.-W. Chang, K.~Lee, and K.~Toutanova, ``{BERT}: Pre-training of
  deep bidirectional transformers for language understanding,'' in
  \emph{NAACL-HLT 2019}, vol. Long and Short Papers, 2019, pp. 4171--4186.
  [Online]. Available: \url{https://aclanthology.org/N19-1423}
\BIBentrySTDinterwordspacing

\bibitem{Sajjad2021NeuronlevelIO}
H.~Sajjad, N.~Durrani, and F.~Dalvi, ``Neuron-level interpretation of deep nlp
  models: A survey,'' \emph{TACL}, vol.~10, pp. 1285--1303, 2021.

\bibitem{Nagamine2015ExploringHD}
T.~Nagamine, M.~L. Seltzer, and N.~Mesgarani, ``Exploring how deep neural
  networks form phonemic categories,'' in \emph{Proc. Interspeech}, 2015.

\bibitem{Belinkov2017AnalyzingHR}
Y.~Belinkov and J.~R. Glass, ``Analyzing hidden representations in end-to-end
  automatic speech recognition systems,'' in \emph{NIPS}, 2017.

\bibitem{Belinkov2019AnalyzingPA}
Y.~Belinkov, A.~Ali, and J.~Glass, ``{Analyzing Phonetic and Graphemic
  Representations in End-to-End Automatic Speech Recognition},'' in \emph{Proc.
  Interspeech 2019}, 2019, pp. 81--85.

\bibitem{deSeyssel2022ProbingPL}
M.~de~Seyssel, M.~Lavechin, Y.~Adi, E.~Dupoux, and G.~Wisniewski, ``Probing
  phoneme, language and speaker information in unsupervised speech
  representations,'' in \emph{Proc. Interspeech}, 2022.

\bibitem{Chrupaa2021VisuallyGM}
G.~Chrupała, ``Visually grounded models of spoken language: A survey of
  datasets, architectures and evaluation techniques,'' \emph{J. Artif. Intell.
  Res.}, vol.~73, pp. 673--707, 2021.

\bibitem{alishahiEncodingPhonologyRecurrent2017}
\BIBentryALTinterwordspacing
A.~Alishahi, M.~Barking, and G.~Chrupala, ``Encoding of phonology in a
  recurrent neural model of grounded speech,'' in \emph{Proc. 21st {{CoNLL}}
  2017}.\hskip 1em plus 0.5em minus 0.4em\relax {ACL}, 2017, pp. 368--378.
  [Online]. Available: \url{https://aclanthology.org/K17-1037}
\BIBentrySTDinterwordspacing

\bibitem{Khorrami2021CanPS}
\BIBentryALTinterwordspacing
K.~Khorrami and O.~Räsänen, ``Can phones, syllables, and words emerge as
  side-products of cross-situational audiovisual learning? - a computational
  investigation,'' \emph{LDR}, vol.~1, no.~1, 2021. [Online]. Available:
  \url{http://ldr.lps.library.cmu.edu/article/id/434/}
\BIBentrySTDinterwordspacing

\bibitem{Singla2022WhatDA}
Y.~K. Singla, J.~Shah, C.~Chen, and R.~R. Shah, ``What do audio transformers
  hear? probing their representations for language delivery \& structure,'' in
  \emph{ICDMW 2022}, 2022, pp. 910--925.

\bibitem{Liu2019MockingjayUS}
A.~T. Liu, S.~wen Yang, P.-H. Chi, P.-C. Hsu, and H.~yi~Lee, ``Mockingjay:
  Unsupervised speech representation learning with deep bidirectional
  transformer encoders,'' \emph{ICASSP 2020}, pp. 6419--6423, 2019.

\bibitem{panayotovLibrispeechASRCorpus2015}
V.~Panayotov, G.~Chen, D.~Povey, and S.~Khudanpur, ``Librispeech: {{An ASR}}
  corpus based on public domain audio books,'' in \emph{ICASSP 2015}, 2015, pp.
  5206--5210.

\bibitem{hsuTextFreeImagetoSpeechSynthesis2021}
\BIBentryALTinterwordspacing
W.-N. Hsu, D.~Harwath, T.~Miller, C.~Song, and J.~Glass, ``Text-{{Free
  Image-to-Speech Synthesis Using Learned Segmental Units}},'' in
  \emph{ACL-IJCNLP 2021}, vol. 1: Long Papers.\hskip 1em plus 0.5em minus
  0.4em\relax {ACL}, 2021, pp. 5284--5300. [Online]. Available:
  \url{https://aclanthology.org/2021.acl-long.411}
\BIBentrySTDinterwordspacing

\bibitem{linMicrosoftCOCOCommon2015}
\BIBentryALTinterwordspacing
T.-Y. Lin, M.~Maire, S.~Belongie, J.~Hays, P.~Perona, D.~Ramanan, P.~Dollar,
  and L.~Zitnick, ``Microsoft {COCO}: Common objects in context,'' in
  \emph{{ECCV}}, 2014. [Online]. Available:
  \url{https://www.microsoft.com/en-us/research/publication/microsoft-coco-common-objects-in-context/}
\BIBentrySTDinterwordspacing

\bibitem{pengSelfSupervisedRepresentationLearning2022}
P.~Peng and D.~Harwath, ``Self-supervised representation learning for speech
  using visual grounding and masked language modeling,'' in \emph{AAAI-SAS
  2022}, 2022.

\bibitem{qiStanzaPythonNatural2020}
\BIBentryALTinterwordspacing
P.~Qi, Y.~Zhang, Y.~Zhang, J.~Bolton, and C.~D. Manning, ``Stanza: {{A Python
  Natural Language Processing Toolkit}} for {{Many Human Languages}},'' in
  \emph{ACL 2020: {{System Demonstrations}}}.\hskip 1em plus 0.5em minus
  0.4em\relax {ACL}, 2020, pp. 101--108. [Online]. Available:
  \url{https://www.aclweb.org/anthology/2020.acl-demos.14}
\BIBentrySTDinterwordspacing

\bibitem{kriegeskorte2008representational}
N.~Kriegeskorte, M.~Mur, and P.~A. Bandettini, ``Representational similarity
  analysis-connecting the branches of systems neuroscience,'' \emph{Frontiers
  in systems neuroscience}, p.~4, 2008.

\bibitem{moschittiMakingTreeKernels2006}
\BIBentryALTinterwordspacing
A.~Moschitti, ``Making {{Tree Kernels Practical}} for {{Natural Language
  Learning}},'' in \emph{EACL 2006}.\hskip 1em plus 0.5em minus 0.4em\relax
  {ACL}, 2006, pp. 113--120. [Online]. Available:
  \url{https://aclanthology.org/E06-1015}
\BIBentrySTDinterwordspacing

\bibitem{collinsConvolutionKernelsNatural2001}
M.~Collins and N.~Duffy, ``Convolution {{Kernels}} for {{Natural Language}},''
  in \emph{NIPS 2001}, vol.~14.\hskip 1em plus 0.5em minus 0.4em\relax {MIT
  Press}, 2001.

\end{thebibliography}
